\documentclass[journal]{IEEEtran} 
\IEEEoverridecommandlockouts
\usepackage{cite}
\usepackage{amsmath,amssymb,amsfonts}
\usepackage{algorithmic}
\usepackage{graphicx}
\usepackage{textcomp}
\usepackage{xcolor}
\usepackage{subcaption}
\usepackage{booktabs}
\usepackage{algorithm}
\usepackage{multirow}
\usepackage{float}
\def\BibTeX{{\rm B\kern-.05em{\sc i\kern-.025em b}\kern-.08em
    T\kern-.1667em\lower.7ex\hbox{E}\kern-.125emX}}
\begin{document}

\title{Single Image Reflection Removal via Self-Supervised Diffusion Models}

\author{
    \IEEEauthorblockN{Zhengyang Lu, Weifan Wang, Tianhao Guo, Feng Wang}
    \\
    \IEEEauthorblockA{\textit{School of Design, Jiangnan University, China}}\\
    
}

\maketitle
	
\begin{abstract}
Reflections often degrade the visual quality of images captured through transparent surfaces, and reflection removal methods suffers from the shortage of paired real-world samples.This paper proposes a hybrid approach that combines cycle-consistency with denoising diffusion probabilistic models (DDPM) to effectively remove reflections from single images without requiring paired training data. The method introduces a Reflective Removal Network (RRN) that leverages DDPMs to model the decomposition process and recover the transmission image, and a Reflective Synthesis Network (RSN) that re-synthesizes the input image using the separated components through a nonlinear attention-based mechanism. Experimental results demonstrate the effectiveness of the proposed method on the SIR$^2$, Flash-Based Reflection Removal (FRR) Dataset, and a newly introduced Museum Reflection Removal (MRR) dataset, showing superior performance compared to state-of-the-art methods.
\end{abstract}
	

\section{Introduction}
Reflections are a common occurrence in images captured through transparent surfaces, such as glass windows, mirrors, or protective covers\cite{blinn1976texture}. These reflections often degrade the visual quality of the captured images, making them less useful for various computer vision and image processing tasks\cite{li2013exploiting,li2020single}. Removing reflections from camera images is a challenging problem that has attracted significant attention in recent years\cite{arvanitopoulos2017single} due to its practical importance in applications such as image enhancement\cite{lu2024self,guo2014robust,lu2022single}, augmented reality \cite{sinha2012image} and computational photography\cite{shih2015reflection,lu2023joint}.

Denoising diffusion probabilistic models (DDPMs) have recently shown remarkable capabilities in modeling complex image distributions and generating high-quality images. Their gradual denoising process is particularly suited for reflection removal as it can progressively separate superimposed image components while maintaining structural coherence. Different from traditional methods that directly predict the reflection-free image, DDPMs can better handle the inherent ambiguity and multimodal nature of the reflection separation problem. Recent studies have demonstrated the effectiveness of combining diffusion models with transformer-based architectures for various image restoration tasks \cite{wang2024coarse,fu2024uncertainty,wang2024improving,lu2022pyramid,dan2024evaluation}, suggesting the potential of leveraging such frameworks for complex degradation scenarios.

\begin{figure}[!t]
	\centering
	\includegraphics[width=\linewidth]{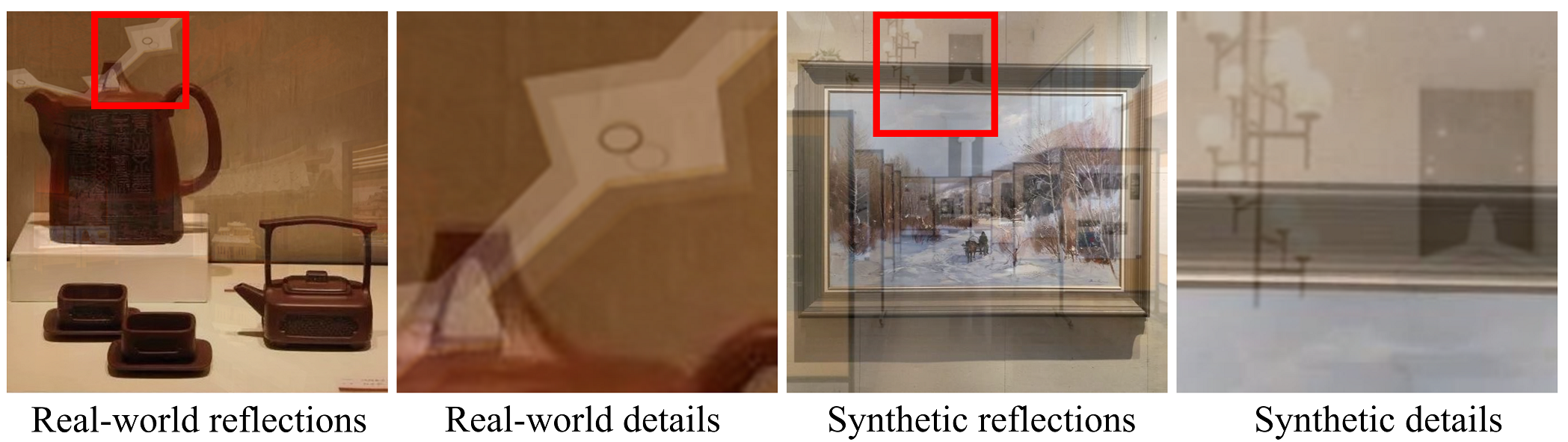}
	\caption{Real-world reflections can be highly complex, often involving multiple superimposed reflections and blurring phenomena. Most methods trained on simple synthetic dataset are struggle to reconstruct transmission images in the wild.}
	\label{fig:prob}
\end{figure}

Existing methods for single image reflection removal (SIRR) can be broadly categorized into two groups: model-based methods and data-driven methods.
Model-based methods \cite{levin2007user,li2014single} typically rely on hand-crafted priors such as gradient sparsity to distinguish transmission from reflection. While these methods can handle simple reflections, they often fail to generalize to real-world images with complex reflections. In contrast, data-driven methods, especially those based on deep learning \cite{fan2017generic,zhang2018single,wei2019single}, learn to separate the layers from training data and have achieved promising results on real images. These methods can handle most reflections by learning from diverse datasets. 
Fig. \ref{fig:prob} illustrates the superimposed reflections from multi-layered media in real-world scenarios. 
Most existing methods trained on synthetic datasets struggle to handle such cases due to the inherent difficulty in simultaneously capturing both reflection and transmission images as training samples.

To address these challenges, we propose a self-supervised approach that combines cycle-consistency with denoising diffusion probabilistic models (DDPM) for single image reflection removal. The proposed method introduces a Reflective Removal Network that leverages DDPMs to model the decomposition process and recover the transmission image, and a Reflective Synthesis Network that re-synthesizes the input image using the separated components through a nonlinear attention-based mechanism.
We conduct extensive experiments on synthetic and real-world datasets, including a newly introduced Museum Reflection Removal (MRR) dataset. The proposed method outperforms state-of-the-art techniques, achieving PSNR gains of 0.50-3.84 dB, SSIM improvements of 0.005-0.074, LPIPS reductions of 0.003-0.057, and RAM decreases of 0.0050-0.0561 compared to the baselines on the SIR$^2$, FRR, and MRR datasets.

The main contributions of this work are:

\begin{itemize}
	\item We propose a reflection removal approach that combines cycle-consistency and DDPMs, effectively modeling the complex distribution of reflections and transmissions.
	\item We introduce an attention-based Reflective Synthesis Network that adaptively fuses the separated components to reconstruct the input image with high fidelity.
	\item We present the Museum Reflection Removal dataset, a diverse collection of real-world and synthetic images with reflections from various artistic contents, facilitating research on reflection removal in challenging scenarios.
	\item We propose a Reflection Artifact Measure to quantify the reflection artifacts in the recovered transmission image, providing a comprehensive assessment of reflection removal performance.
	\item We conduct extensive experiments and ablation studies, demonstrating the superiority of the proposed method over state-of-the-art techniques on multiple datasets.
\end{itemize}

\section{Related Work}

Reflection removal from images has been an active area in computer vision. Existing approaches can be broadly categorized into two main groups: model-based methods and data-driven methods.

\subsection{Model-Based SIRR Methods}

Model-based methods for single image reflection removal typically rely on handcrafted priors and optimization techniques to separate the reflection and transmission layers. Levin and Weiss \cite{levin2007user} proposed a user-assisted approach that requires manual labeling of gradients to separate the layers. Li and Brown \cite{li2014single} developed a method based on optimizing a Laplacian data fidelity term and a gradient sparsity prior to remove reflections. These methods often struggle to handle complex reflections and require manual intervention, limiting their practicality.

Several pioneering approaches have laid the foundation for model-based reflection removal. Farid and Adelson \cite{farid1999separating} introduced independent components analysis to separate reflections and lighting, demonstrating the potential of statistical methods in layer decomposition. Building upon this, Szeliski et al. \cite{szeliski2000layer} developed an optimal approach for recovering layer images through constrained least squares and iterative refinement, though their method required multiple input images. A significant advancement came from Levin et al. \cite{levin2004separating}, who proposed using local features and belief propagation to decompose a single image by minimizing the total amount of edges and corners. Kong et al. \cite{kong2011high} later enhanced the separation quality by leveraging polarized images, introducing a constrained optimization framework that exploits mutually exclusive image information. While these model-based methods established important theoretical foundations and demonstrated promising results in controlled scenarios, they often struggle with complex real-world reflections due to their reliance on simplified assumptions about image statistics and gradient distributions.

\subsection{Data-Driven Methods}

More recently, deep learning-based approaches have gained popularity for single-image reflection removal. Fan et al. \cite{fan2017generic} proposed a deep neural network architecture that learns to suppress reflections by explicitly modeling the ghosting effects. Zhang et al. \cite{zhang2018single} introduced a perceptual loss function and a reflection removal network to recover the background layer from a single image. Wei et al. \cite{wei2019single} proposed an edge-guided reflection removal network that utilizes edge information to improve the separation of transmission and reflection layers. Yang et al. \cite{yang2018seeing} developed a bidirectional deep network with a recurrent structure to iteratively refine the reflection removal results.

Several works have also explored the use of generative adversarial networks (GANs) for single-image reflection removal. Wan et al. \cite{wan2018crrn} proposed a concurrent reflection removal network (CRRN) that utilizes a transformation-induced image formation model and a concurrent optimization strategy to remove reflections. Wen et al. \cite{wen2019single} developed a dual attention network that exploits channel and spatial attention mechanisms to effectively remove reflections. Liu and Lu \cite{liu2020separate} propose an unsupervised approach using GANs with self-supervision and cycle consistency constraints. Abiko and Ikehara \cite{abiko2019single} introduce a gradient constraint loss in GAN framework to minimize the correlation between background and reflection layers.

To address the challenge of limited paired training data, recent works have explored cycle consistency and physically-based approaches.RahmaniKhezri et al. \cite{rahmanikhezri2022unsupervised} develop an unsupervised method using cross-coupled deep networks that leverages semantic features for layer separation, demonstrating strong performance without requiring extensive paired datasets. Kim et al. \cite{kim2020single} utilize physically-based rendering to synthesize training data, successfully reproducing spatially variant anisotropic effects of glass reflection. They further introduce a backtrack network for removing complicated ghosting and defocused effects. These methods demonstrate the importance of realistic training data synthesis and cycle-consistent learning in improving reflection removal performance.

Multiple-image reflection removal methods utilize additional images or priors to aid the reflection removal process. These methods typically require capturing multiple images of the same scene under different conditions, such as varying polarization states or focus settings. Schechner et al. \cite{schechner2000polarization} proposed a method that uses a sequence of images captured with different polarization filters to separate the reflection and transmission layers. Agrawal et al. \cite{agrawal2005removing} utilized a pair of flash and no-flash images to remove reflections by exploiting the differences in the reflective properties of the two images. Kong et al. \cite{kong2013physically} proposed a method that uses multiple images captured with different focus settings to remove reflections based on depth-of-field differences. Xue et al. \cite{xue2015computational} developed a computational approach that utilizes a pair of images captured from slightly different viewpoints to remove reflections by exploiting the motion parallax.

More recently, deep learning-based approaches have also been explored for multiple-image reflection removal. Fan et al. \cite{fan2017generic} proposed a deep architecture that learns to remove reflections from a pair of images captured under different polarization states. Li et al. \cite{li2019single} developed a deep neural network that utilizes a pair of images captured with different focus settings to remove reflections. Li et al. \cite{li2019feedback} introduced a deep learning framework that uses a sequence of images captured with different polarization angles to remove reflections and recover the underlying scene. Wieschollek et al. \cite{wieschollek2018separating} proposed a kernel-based method that separates the reflection and transmission layers using a kernel estimation approach.

Despite the progress made by these data-driven methods, they often struggle to handle strong and complex reflections, resulting in artifacts or residual reflections in the recovered images. Our proposed approach aims to address these limitations by leveraging the power of cycle-consistency and denoising diffusion probabilistic models to effectively remove reflections from single images.

\section{Proposed Method}

\begin{figure*}[!t]
	\centering
	\includegraphics[width=\linewidth]{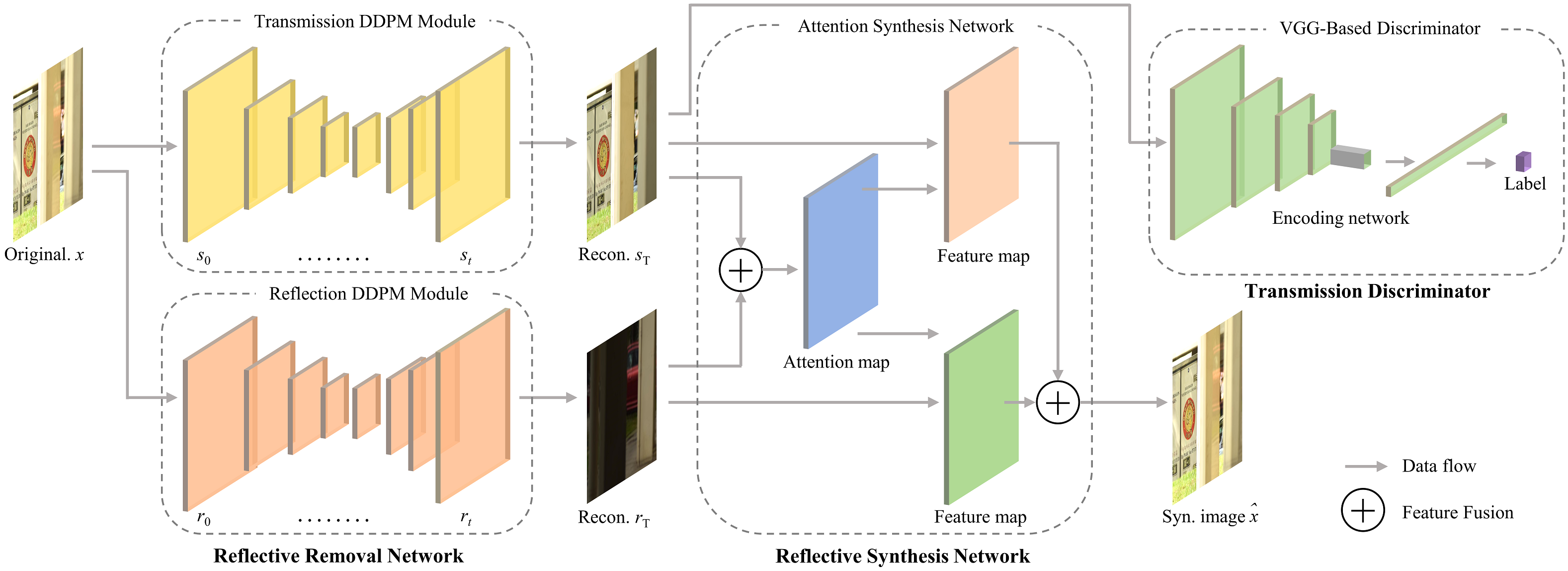}
	\caption{The self-supervised diffusion model has three main components: Reflective Removal Network, Reflective Synthesis Network, and Transmission Discriminator.}
	\label{fig:arc}
\end{figure*}

In this section, we present a self-supervised method for single image reflection removal, which combines cycle-consistency and denoising diffusion probabilistic models (DDPMs). As shown in Fig.\ref{fig:arc}, the proposed approach consists of three main components: a Reflective Removal Network (RRN), a Reflective Synthesis Network (RSN) and a Transmission Discriminator (TD). The RRN utilizes the DDPM framework to model the decomposition process and recover the transmission image from the input image, while the RSN synthesizes the input image with reflections by combining the recovered transmission and reflection components through a nonlinear attention-based mechanism. The cycle-consistent framework enables the learning of mapping functions between the domains without paired training data.

\subsection{Cycle-Consistent Framework}

The cycle-consistent framework enables the learning of mapping functions between two domains without paired training data. Let $\mathcal{X}$ denote the domain of camera images containing reflections, and $\mathcal{S}$ and $\mathcal{R}$ represent the domains of the transmission images and reflection images, respectively. We define two mapping functions: $G: \mathcal{X} \rightarrow \mathcal{S} \times \mathcal{R}$ and $F: \mathcal{S} \times \mathcal{R} \rightarrow \mathcal{X}$. The function $G$ aims to decompose an input camera image $x \in \mathcal{X}$ into its transmission image component $s \in \mathcal{S}$ and reflection image component $r \in \mathcal{R}$, while $F$ synthesizes the camera image $x \in \mathcal{X}$ from the recovered transmission image $s$ and reflection image $r$.

The cycle-consistency constraint enforces that the mapping functions $G$ and $F$ should be bijective, ensuring that the recovered components can be reconstructed back to the original input image. Mathematically, this constraint can be expressed as:

\begin{equation}
	x \approx F(G(x)) \quad \forall x \in \mathcal{X}
\end{equation}

Additionally, we introduce an inverse mapping $G^{-1}: \mathcal{S} \times \mathcal{R} \rightarrow \mathcal{X}$, which synthesizes the camera image directly from the transmission and reflection image components. The inverse cycle-consistency constraint is then formulated as:

\begin{equation}
	s, r \approx G(G^{-1}(s, r)) \quad \forall s \in \mathcal{S}, r \in \mathcal{R}
\end{equation}

\subsection{Reflective Removal Network}

The Reflective Removal Network (RRN) is responsible for decomposing the input camera image into its transmission image and reflection image components. The proposed approach builds upon the denoising diffusion probabilistic model (DDPM) framework\cite{ho2020denoising} to effectively model the decomposition process and handle complex reflections.

\begin{figure}[!t]
	\centering
	\includegraphics[width=0.95\linewidth]{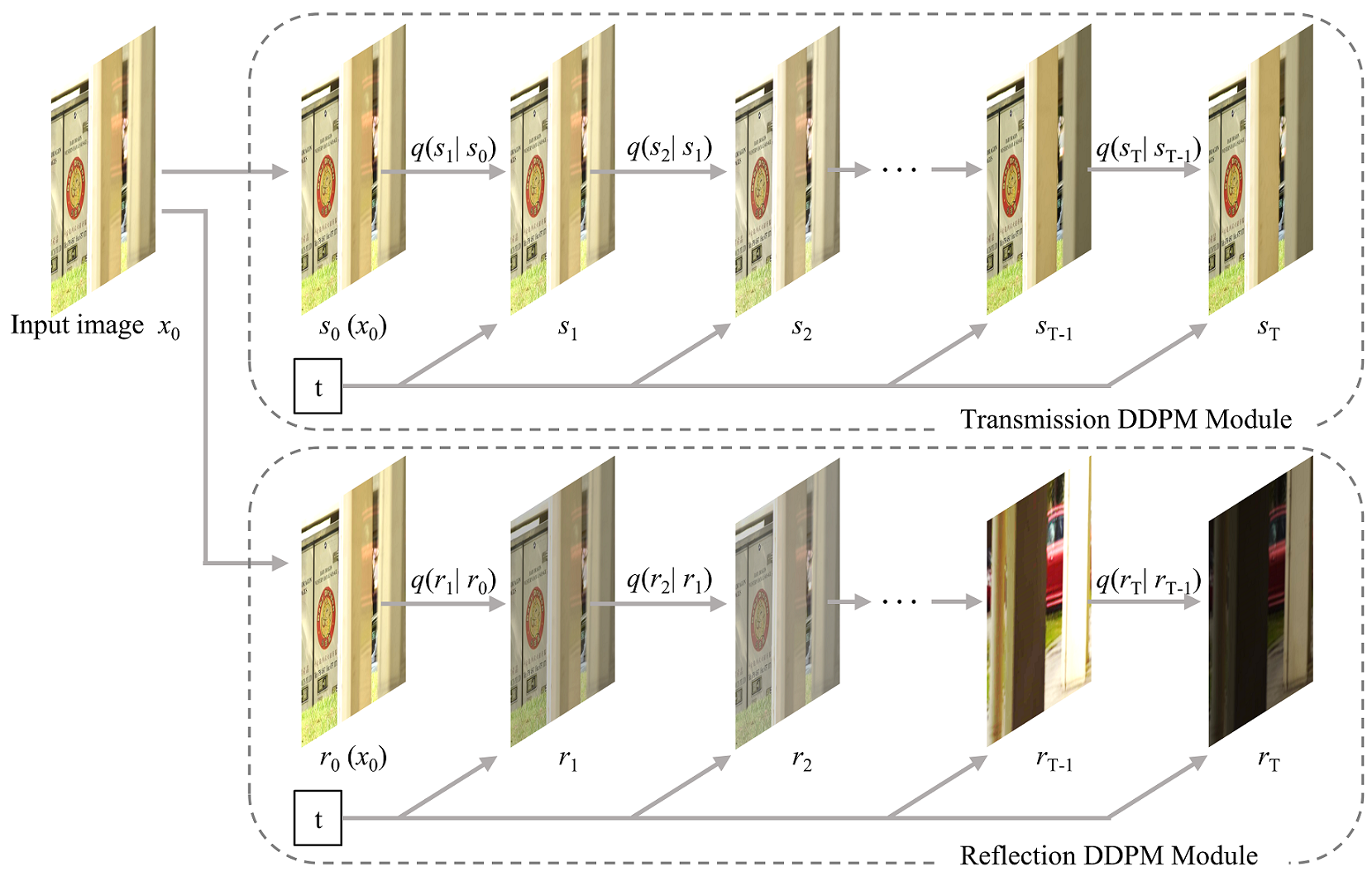}
	\caption{Dual DDPM ensures accurate transmission and reflection image predictions.}
	\label{fig:ddpmarc}
\end{figure}

The DDPM framework learns a mapping from a noisy image distribution to the clean image distribution through a forward diffusion process and a reverse diffusion process. 
Let $x_0 \in \mathbb{R}^{H \times W \times 3}$ denote the input camera image, and $s_0 \in \mathbb{R}^{H \times W \times 3}$ and $r_0 \in \mathbb{R}^{H \times W \times 3}$ represent the corresponding ground truth transmission image and reflection image, respectively, as shown in Fig.\ref{fig:ddpmarc}. The DDPM framework defines a forward diffusion process $q(x_t | x_{t-1})$ that gradually adds Gaussian noise to the input image over $T$ timesteps, resulting in a sequence of noisy images $x_{t=1}^T$. The forward diffusion process can be described as a Markov chain with the following transition probability:

\begin{equation}
	q(x_t | x_{t-1}) = \mathcal{N}(x_t; \sqrt{1 - \beta_t} x_{t-1}, \beta_t I),
\end{equation}
where $\beta_t \in (0, 1)$ is a variance schedule that determines the amount of noise added at each timestep.

In the context of reflection removal, we extend the forward diffusion process to generate two separate noisy image sequences: one for the transmission component ${s}_{t=1}^T$ and another for the reflection component ${r}_{t=1}^T$. This extension allows the RRN to model the decomposition of the input camera image into its constituent transmission and reflection components. The forward diffusion processes for the transmission and reflection components are defined as follows:
\begin{align}
	q(s_t | s_{t-1}) &= \mathcal{N}(s_t; \sqrt{1 - \beta_t} s_{t-1}, \beta_t I), \\
	q(r_t | r_{t-1}) &= \mathcal{N}(r_t; \sqrt{1 - \beta_t} r_{t-1}, \beta_t I).
\end{align}
where $s_t$ and $r_t$ denote the noisy transmission and reflection images at timestep $t$, respectively.

The objective of the Reflective Removal Network is to learn the reverse diffusion processes $p_\theta(s_{t-1} | s_t)$ and $p_\theta(r_{t-1} | r_t)$ that recursively denoise the noisy transmission and reflection images, ultimately reconstructing the clean transmission and reflection images $s_0$ and $r_0$, respectively. By leveraging the DDPM framework, the RRN can effectively model the complex distribution of reflections and transmissions in camera images, enabling accurate decomposition and removal of reflections.

Following the DDPM framework, the reverse diffusion processes are parameterized by a shared neural network with parameters $\theta$, denoted as $\epsilon_\theta(\cdot, t)$. The reverse diffusion processes can be described as follows:
\begin{align}
	p_\theta(s_{t-1} | s_t) &= \mathcal{N}(s_{t-1}; \mu_\theta(s_t, t), \sigma_t^2 I), \\
	p_\theta(r_{t-1} | r_t) &= \mathcal{N}(r_{t-1}; \mu_\theta(r_t, t), \sigma_t^2 I),
\end{align}
where $\mu_\theta(\cdot, t) = \frac{1}{\sqrt{1 - \beta_t}} \left(z_t - \frac{\beta_t}{\sqrt{1 - \bar{\alpha}_t}} \epsilon_\theta(z_t, t) \right)$, $\sigma_t^2 = \beta_t$, and $\bar{\alpha}_t = \prod_{i=1}^t (1 - \beta_i)$.

The training objective for the Reflective Removal Network is to minimize the combination of the denoising loss and the reconstruction loss. The denoising loss $\mathcal{L}_\text{de}$ is derived from the variational lower bound of the negative log-likelihood of the data distribution. Specifically, for the transmission component $s_0$, the denoising loss $\mathcal{L}_\text{de}$ can be expressed as:
\begin{align}
	\mathcal{L}_\text{de}(s_0) &= \mathbb{E}{s_0, \epsilon} \left[ \frac{1}{T} \sum_{t=1}^T \left| \epsilon - \epsilon_\theta(\sqrt{\bar{\alpha}_t} s_0 + \sqrt{1 - \bar{\alpha}_t} \epsilon, t) \right|_2^2 \right] \\
	&= \mathbb{E}{s_0, \epsilon} \left[ \frac{1}{T} \sum_{t=1}^T \left| \epsilon - \epsilon_\theta(s_t, t) \right|_2^2 \right],
\end{align}
where $\epsilon \sim \mathcal{N}(0, I)$ is the standard Gaussian noise, and $s_t = \sqrt{\bar{\alpha}_t} s_0 + \sqrt{1 - \bar{\alpha}_t} \epsilon$ is the noisy transmission image at timestep $t$. The denoising loss measures the difference between the predicted noise $\epsilon_\theta(s_t, t)$ and the actual noise $\epsilon$ used to generate the noisy image $s_t$. By minimizing this loss, the network learns to predict the noise that needs to be removed from the noisy image to reconstruct the clean transmission image.

Similarly, for the reflection component $r_0$, the denoising loss can be expressed as:
\begin{align}
	\mathcal{L}_\text{de}(r_0) &= \mathbb{E}{r_0, \epsilon} \left[ \frac{1}{T} \sum_{t=1}^T \left| \epsilon - \epsilon_\theta(\sqrt{\bar{\alpha}_t} r_0 + \sqrt{1 - \bar{\alpha}_t} \epsilon, t) \right|_2^2 \right] \\
	&= \mathbb{E}{r_0, \epsilon} \left[ \frac{1}{T} \sum_{t=1}^T \left| \epsilon - \epsilon_\theta(r_t, t) \right|_2^2 \right],
\end{align}
where $r_t = \sqrt{\bar{\alpha}_t} r_0 + \sqrt{1 - \bar{\alpha}_t} \epsilon$ is the noisy reflection image at timestep $t$.

The overall denoising loss for the Reflective Removal Network is the sum of the denoising losses for the transmission and reflection components:
\begin{equation}
	\mathcal{L}_\text{de} = \mathcal{L}_\text{de}(s_0) + \mathcal{L}_\text{de}(r_0).
\end{equation}

In addition to the denoising loss, we introduce a reconstruction loss to ensure the fidelity of the reconstructed transmission and reflection images:
\begin{equation}
	\mathcal{L}_\text{rec} = \mathbb{E}_{s_0, r_0} \left[ \left| s_0 - \hat{s}_0 \right|_1 + \left| r_0 - \hat{r}_0 \right|_1 \right],
\end{equation}
where $\hat{s}_0$ and $\hat{r}_0$ are the reconstructed transmission and reflection images, respectively, obtained by recursively applying the reverse diffusion processes starting from the noisy images at timestep $T$.

\subsection{Transmission Discriminator}

To address potential network degradation during unpaired training with the cycle-consistent network and to improve the accuracy of the recovered transmission image, we introduce a Transmission Discriminator (TD) as an adversarial component in the proposed framework.

The Transmission Discriminator is a convolutional neural network that learns to distinguish between the recovered transmission images generated by the Reflective Removal Network (RRN) and the real transmission images from the training dataset. By incorporating this discriminator, we can ensure that the RRN generates visually realistic and accurate transmission images.
The Transmission Discriminator aims to minimize the adversarial loss, which is defined as:
\begin{align}
	\mathcal{L}_\text{adv} = \mathbb{E}&{s \sim p_\text{data}(s)}[\log D(s)] +\\ &\mathbb{E}{x \sim p_\text{data}(x)}[\log(1 - D(G(x)))],
\end{align}
where $D$ denotes the Transmission Discriminator, $G$ represents the Reflective Removal Network, $s$ is a real transmission image sampled from the data distribution $p_\text{data}(s)$, and $x$ is an input camera image sampled from the data distribution $p_\text{data}(x)$.

The RRN is trained to minimize the adversarial loss, encouraging it to generate transmission images that are indistinguishable from real ones. The Transmission Discriminator, on the other hand, is trained to maximize the adversarial loss, improving its ability to distinguish between real and generated transmission images. In the full framework, the TD enhances the quality of the recovered transmission images by forcing the RRN to generate more realistic and accurate results. This adversarial training process helps maintain the complexity of the network and prevents degradation during unpaired training.

The overall training objective for the Reflective Removal Network is updated to include the adversarial loss:
\begin{equation}
	\mathcal{L}_\text{RRN} = \lambda_1 \mathcal{L}_\text{de} + \lambda_2 \mathcal{L}_\text{rec} + \lambda_3 \mathcal{L}_\text{adv},
\end{equation}
where $\lambda_1$, $\lambda_2$ and $\lambda_3$ are hyper-parameters controlling the relative importance of each loss term.

By incorporating the Transmission Discriminator, the proposed framework benefits from the adversarial training process, which helps maintain the complexity of the network and improves the accuracy of the recovered transmission images.

\subsection{Reflective Synthesis Network}

The Reflective Synthesis Network (RSN) aims to synthesize the input image with reflections by combining the recovered transmission and reflection components. We propose a nonlinear attention-based combination module to adaptively fuse the transmission and reflection components while preserving the details and structures of the input image.

\begin{figure}[!t]
	\centering
	\includegraphics[width=\linewidth]{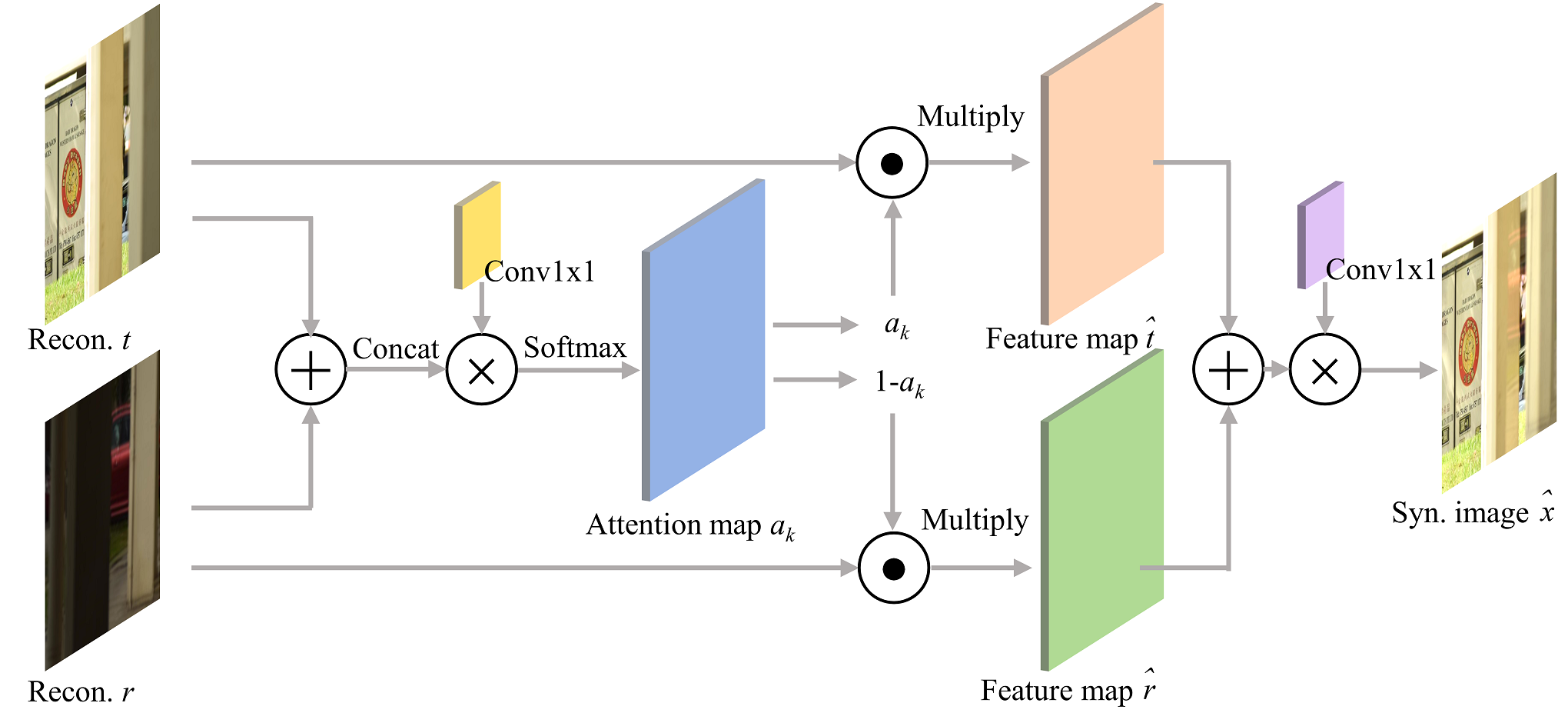}
	\caption{Reflective synthesis network restores the original image from transmission and reflection image with attention mechanism.}
	\label{fig:rsn}
\end{figure}

As shown in Fig.\ref{fig:rsn}, the RSN aims to synthesize the input image $\hat{x} \in \mathcal{X}$ by applying a nonlinear superposition of $s$ and $r$, guided by an attention mechanism.
We first compute a set of spatial attention maps $A = {a_1, a_2, \ldots, a_K}$, where $K$ is the number of attention heads. Each attention map $a_k \in \mathbb{R}^{H \times W}$ is obtained by applying a convolutional operation followed by a softmax activation to the concatenation of $s$ and $r$:

\begin{equation}
	a_k = \text{softmax}(f_c([s, r]))
\end{equation}
where $f_c$ represents a convolutional neural network with learnable parameters, and $[\cdot, \cdot]$ denotes the concatenation operation along the channel dimension.

The attention maps $A$ are then used to modulate the transmission image $t$ and the reflection image $s$, producing attended feature maps $\hat{s}$ and $\hat{r}$, respectively:

\begin{align}
	\hat{s} &= \sum_{k=1}^K a_k \odot s \\
	\hat{r} &= \sum_{k=1}^K (1 - a_k) \odot r
\end{align}

where $\odot$ denotes the element-wise multiplication operation.

The final synthesized camera image $\hat{x}$ is obtained by combining the attended feature maps $\hat{t}$ and $\hat{r}$ through a nonlinear transformation:

\begin{equation}
	\hat{x} = g([\hat{s}, \hat{r}])
\end{equation}

where $g$ is a convolutional neural network that learns to synthesize the camera image with reflections from the attended transmission and reflection image components.

To ensure that the synthesized camera image closely resembles the original input image, we introduce a cycle-consistency loss as the overall objective function for the Reflective Synthesis Network:

\begin{align}
	\mathcal{L}_\text{RSN} = \mathcal{L}_\text{cyc} = \left\lVert x - \hat{x} \right\rVert_1 =\left\lVert x - F(G(x)) \right\rVert_1
\end{align}

\subsection{Training and Inference}

The training and inference processes of the proposed method involve two main components: the complete data training flow and the single input inference flow in the cycle-consistent network.

\subsubsection{Paired Sample Training Flow}
The complete data training flow leverages paired data, consisting of input camera images with reflections and their corresponding ground truth transmission images. The training process alternates between updating the Reflective Removal Network (RRN) and the Reflective Synthesis Network (RSN). Algorithm~\ref{alg:training_flow} outlines the paired sample training flow.

\begin{algorithm}[!t]
	\caption{Paired Sample Training Flow}
	\label{alg:training_flow}
	\begin{algorithmic}[1]
		\REQUIRE Paired training data $\{(x_i, s_i, r_i)\}_{i=1}^N$, where $x_i$ is the input camera image, $s_i$ is the ground truth transmission image and $r_i$ is the ground truth reflection image
		\ENSURE Trained RRN and RSN models
		\WHILE{not converged}
		\STATE Sample a mini-batch of paired data $\{(x, s, r)\}$
		\STATE \textbf{Update RRN:}
		\STATE \hspace{1em} Compute $\hat{s}, \hat{r} = \text{RRN}(x)$
		\STATE \hspace{1em} Compute denoising loss $\mathcal{L}_\text{de}$
		\STATE \hspace{1em} Compute reconstruction loss $\mathcal{L}_\text{rec}$
		\STATE \hspace{1em} Compute adversarial loss $\mathcal{L}_\text{adv}$ 
		\STATE \hspace{1em} Update RRN parameters to minimize \\
		\hspace{3em} $\mathcal{L}_\text{RRN} = \lambda_1 \mathcal{L}_\text{de} + \lambda_2 \mathcal{L}_\text{rec} + \lambda_3 \mathcal{L}_\text{adv}$
		\STATE \hspace{1em} Update TD parameters to minimize $\mathcal{L}_\text{adv}$
		\STATE \textbf{Update RSN:}
		\STATE \hspace{1em} Compute $\hat{x} = \text{RSN}(\hat{s}, \hat{r})$
		\STATE \hspace{1em} Compute cycle-consistency loss $\mathcal{L}_\text{cyc}$ 
		\STATE \hspace{1em} Update RSN parameters to minimize $\mathcal{L}_\text{cyc}$
		\ENDWHILE
	\end{algorithmic}
\end{algorithm}

During the RRN update step, the input camera image $x$ is passed through the RRN to obtain the estimated transmission image $\hat{s}$ and reflection image $\hat{r}$. The denoising loss $\mathcal{L}_\text{de}$ and reconstruction loss $\mathcal{L}_\text{rec}$ are computed based on the estimated and ground truth transmission images. The RRN parameters are then updated using gradient descent to minimize these losses.
In the RSN update step, the estimated transmission image $\hat{s}$ and reflection image $\hat{r}$ are fed into the RSN to synthesize the camera image $\hat{x}$. The cycle-consistency loss $\mathcal{L}_\text{cyc}$ is computed between the input camera image $x$ and the synthesized image $\hat{x}$. The RSN parameters are updated using gradient descent to minimize the cycle-consistency loss.

\subsubsection{Unpaired Sample Training Flow}
During inference, The proposed method takes a single input camera image with reflections and aims to remove the reflections to obtain the transmission image. The single input inference flow utilizes the trained RRN, RSN, and TD models in a cycle-consistent framework. Algorithm~\ref{alg:inference_flow_updated} describes the unpaired sample training flow with the transmission discriminator.

\begin{algorithm}[!t]
	\caption{Unpaired Sample Training Flow}
	\label{alg:inference_flow_updated}
	\begin{algorithmic}[1]
		\REQUIRE Unpaired training data ${x}_{i=1}^N$, where $x_i$ is the input camera image, $s_i$ is the transmission image and $r_i$ is the reflection image
		\ENSURE Trained RRN and RSN models
		\WHILE{not converged}
		\STATE Sample a mini-batch of unpaired data ${x}$ and ${s}$
		\STATE \textbf{Update RRN, RSN, and TD:}
		\STATE \hspace{1em} Compute $\hat{s}, \hat{r} = \text{RRN}(x)$
		\STATE \hspace{1em} Compute $\hat{x} = \text{RSN}(\hat{s}, \hat{r})$
		\STATE \hspace{1em} Compute cycle-consistency loss \\
		\hspace{3em}	$\mathcal{L}_\text{cyc}=\left\lVert x - \hat{x} \right\rVert_1$
		\STATE \hspace{1em} Compute adversarial loss \\
		\hspace{3em}	$\mathcal{L}_\text{adv}=\text{TD}(\hat{s})$
		\STATE \hspace{1em} Update RRN and RSN parameters to minimize\\
		\hspace{3em} $\mathcal{L}_\text{cyc} + \lambda_\text{adv}\mathcal{L}_\text{adv}$
		\ENDWHILE
	\end{algorithmic}
\end{algorithm}

Given an input camera image $x$, the RRN first predicts the transmission image $\hat{s}$ and reflection image $\hat{r}$. The RSN then takes $\hat{s}$ and $\hat{r}$ as inputs and synthesizes the reconstructed camera image $\hat{x}$. The cycle-consistency loss $\mathcal{L}_\text{cyc}$ is computed between the input image $x$ and the reconstructed image $\hat{x}$ to enforce the bijective mapping between the domains.
In addition to the cycle-consistency loss, we also incorporates the TD to improve the quality of the predicted transmission image $\hat{s}$. The adversarial loss $\mathcal{L}_\text{adv}$ is computed by passing $\hat{s}$ through the TD, which learns to distinguish between real and predicted transmission images.
During training, the RRN and RSN parameters are updated simultaneously to minimize the combined loss function, which combines cycle-consistency loss and adversarial loss.

\subsubsection{Model Inference}

During inference, our method takes a single input image and processes it through the trained RRN to obtain reflection-free results. This process is deterministic and requires no iterative optimization, enabling efficient real-world applications.

The inference pipeline consists of two stages: first, the RRN decomposes the input into transmission and reflection components through a series of denoising steps. Then, only the transmission component is extracted as the final output, discarding the reflection component.

Computationally, inference requires only a single forward pass through the RRN, making it significantly more efficient than training. The RSN and TD components are not used during inference as they serve only training purposes. For an input image of resolution $H\times W$, the inference process has a computational complexity of $O(HW)$, with actual runtime primarily determined by the network architecture and available computing resources.

\begin{algorithm}[t]
	\caption{Model Inference Process}
	\begin{algorithmic}[1]
		\REQUIRE Input image x, Trained RRN model
		\ENSURE Reflection-free transmission image
		\STATE Load trained RRN parameters
		\STATE $s, r = \text{RRN}(x)$ \COMMENT{Decompose input}
		\STATE $\hat{s} = \text{normalize}(s)$ \COMMENT{Post-process transmission}
		\RETURN $\hat{s}$ \COMMENT{Return transmission only}
	\end{algorithmic}
\end{algorithm}

\section{Experimental Results}
\label{sec:experiments}

In this section, we present comprehensive evaluations of the proposed method for reflective removal from camera images. We first introduce the datasets used for training and testing, followed by details of the experimental setup. We then conduct ablation studies to analyze the contribution of different components and loss functions. Furthermore, we compare the proposed method with existing state-of-the-art techniques and demonstrate its effectiveness on real-world scenarios.

\subsection{Datasets}

\begin{figure}[tbp]
	\centering
	\includegraphics[width=\linewidth]{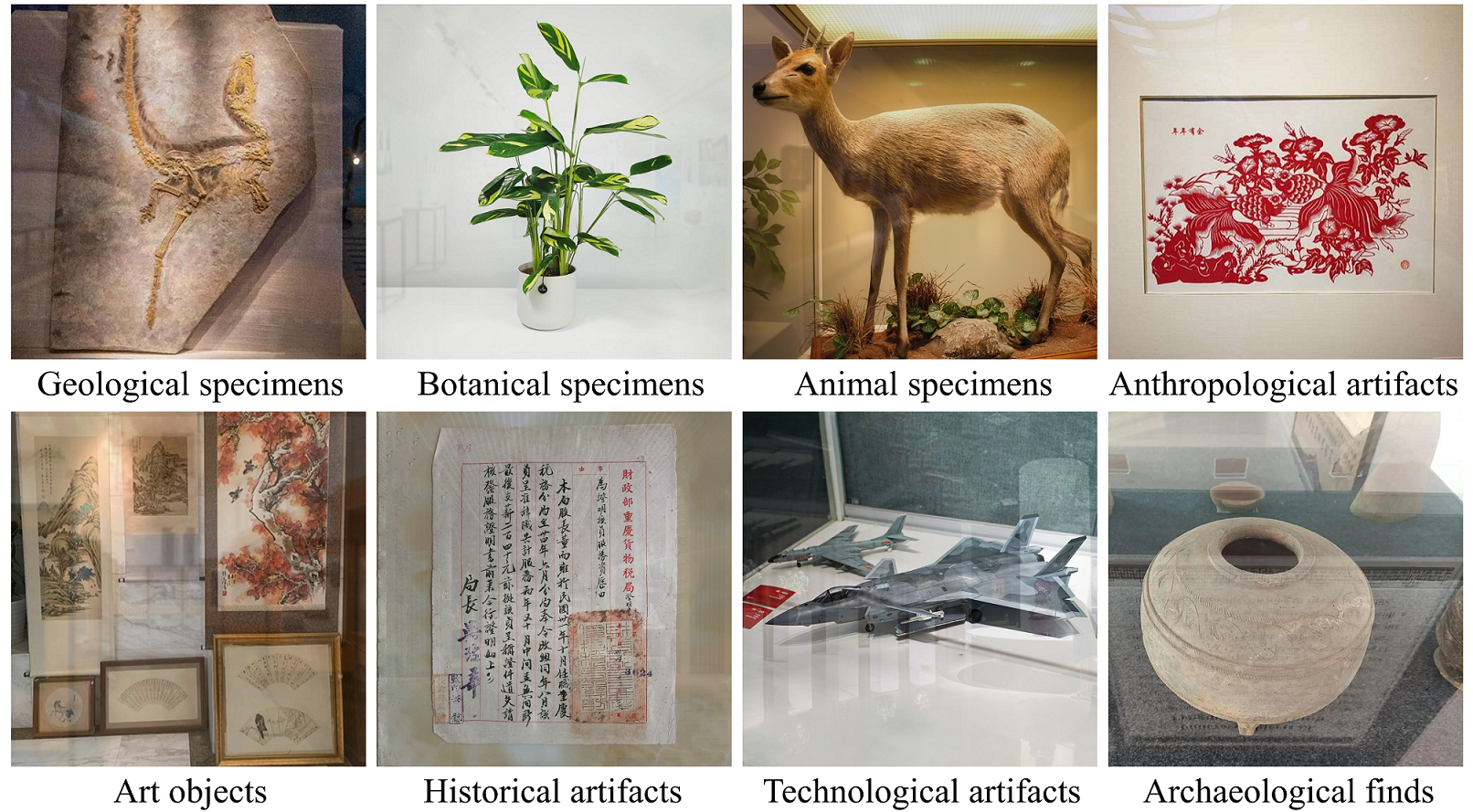}
	\caption{The Museum Reflection Removal Dataset includes exhibition samples with reflections from various fields.}
	\label{fig:datasetdemo}
\end{figure}

\begin{figure}[tbp]
	\centering
	\includegraphics[width=\linewidth]{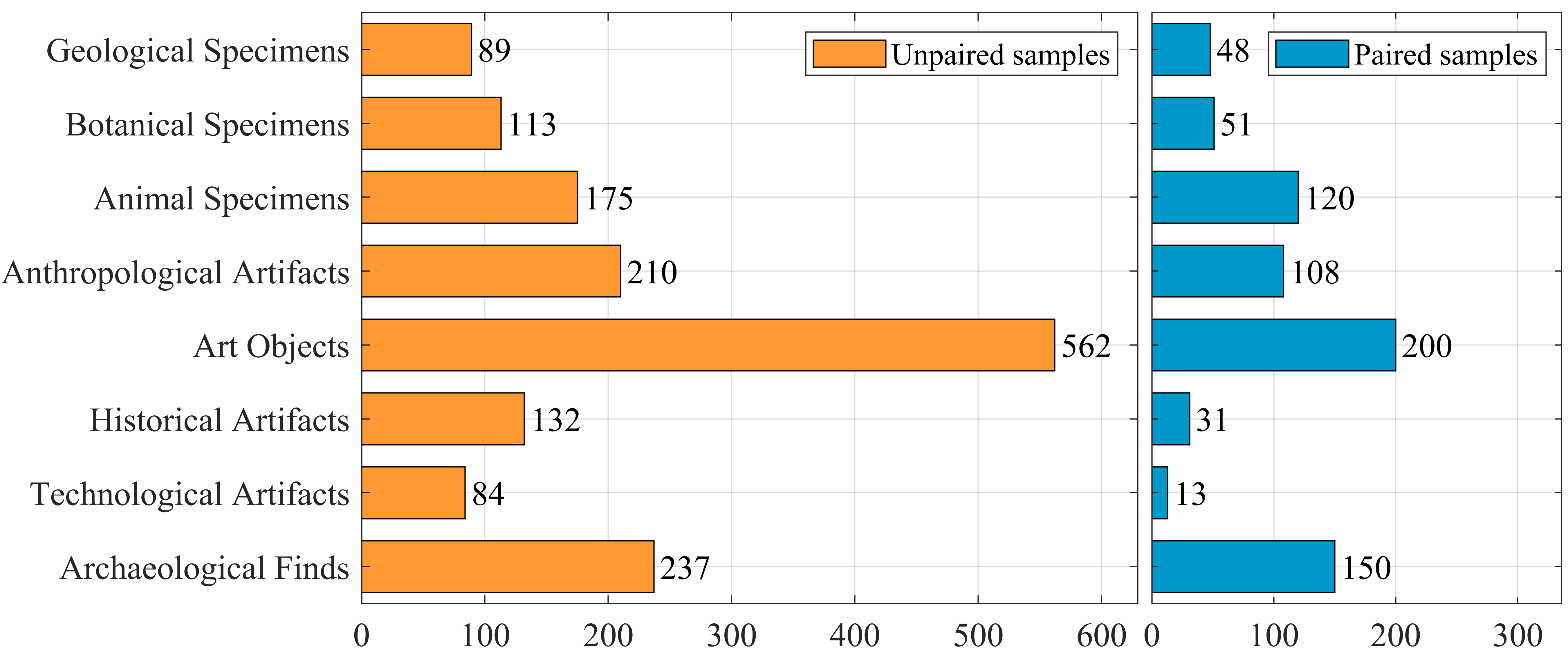}
	\caption{Artifacts Distribution in MRR Dataset.}
	\label{fig:datadistribution}
\end{figure}

We evaluate the proposed method on real-world datasets and Museum Reflection Removal (MRR) datasets:

\paragraph{SIR$^2$ Dataset} For real-world evaluation, we utilize publicly available SIR$^2$ dataset \cite{wan2017benchmarking}. The SIR$^2$ dataset contains 500 real-world image pairs captured through glass surfaces with reflections. We use SIR$^2$ dataset to assess the generalization ability of existing methods on real-world scenarios.
\paragraph{Flash-Based Reflection Removal (FRR) Dataset} The FRR dataset\cite{lei2021robust} is the first to provide RAW flash/no-flash image pairs for reflection removal. It contains 157 real-world scenes captured by Nikon Z6 and Huawei Mate30, and 1964 synthetic scenes. Each real-world scene includes ambient images with and without reflection captured under flash and no-flash conditions, and a no-flash reflection image. The dataset enables flash-based reflection removal.
\paragraph{Museum Reflection Removal (MRR) Dataset} To evaluate the performance of the proposed method on diverse artistic contents in museums, we introduce the Museum Reflection Removal (MRR) dataset. As shown in Fig.\ref{fig:datadistribution}, the proposed dataset collected 1,621 high-quality unpaired and 721 paired images from various museums and exhibitions, comprising geological specimens, botanical specimens, animal specimens, anthropological artifacts, art objects, historical artifacts, technological artifacts, and archaeological finds. As shown in Fig.\ref{fig:datasetdemo}, each image in the dataset contains reflections caused by protective glass or display cases, presenting a challenging real-world scenario for reflection removal algorithms. To facilitate paired training, the paired images are achieved by combining reflection-free samples with exhibition scenes to create realistic reflections.

\subsection{Experimental Setup}

We implement the proposed method using the PyTorch deep-learning framework \cite{paszke2019pytorch}. All experiments were conducted on a workstation with NVIDIA GeForce RTX 4090s GPUs, 128GB RAM, and Intel Xeon Gold 6226R CPU. The model training takes approximately 48 hours for both stages combined.
The RRN and the Reflective Synthesis Network (RSN) are trained jointly using the Adam optimizer \cite{kingma2014adam} with a learning rate of 1e-4 and a batch size of 8. The RRN is based on the DDPM architecture, with a modified UNet backbone\cite{ronneberger2015u} consisting of 8 downsampling and 8 upsampling layers. The RSN employs a similar architecture with 4 downsampling and 4 upsampling layers, along with the attention-based combination module. The hyperparameters $\lambda_1$, $\lambda_2$, and $\lambda_3$, which control the weights of the denoising loss, reconstruction loss, and adversarial loss, respectively, are set to 1.0, 10.0, and 0.1 based on empirical observations.

\begin{table}[!t]
	\centering
	\caption{Network Architecture and Training Parameters}
	\label{tab:network_params}
	\begin{tabular}{lll}
		\toprule
		Component & Parameter & Value \\
		\midrule
		\multirow{5}{*}{RRN} & Number of layers & 8 down + 8 up \\
		& Channel dimensions & [128,256,512,512,512,512,512,512] \\
		& Diffusion timesteps & T = 1000 \\
		& Beta schedule & Linear ($10^{-4}$ to 0.02) \\
		& Attention resolutions & 16$\times$16, 32$\times$32 \\
		\midrule
		\multirow{6}{*}{RSN} & Feature extraction & 4 down + 4 up \\
		& Channel dimensions & [64,128,256,512] \\
		& Attention module & Single-layer with 1$\times$1 conv \\
		& Softmax normalization & Channel-wise \\
		& Feature fusion & Dual-stream weighted sum \\
		& Final synthesis & 1$\times$1 conv \\
		\bottomrule
	\end{tabular}
\end{table}

Table \ref{tab:network_params} presents the detailed architecture and training parameters of our networks. The RRN adopts a U-Net backbone with symmetric skip connections, where the number of channels doubles after each downsampling operation until reaching 512. For the RSN, we implement a novel attention-based fusion mechanism that operates on the reconstructed transmission $s$ and reflection $r$ images. The attention module uses 1$\times$1 convolutions followed by softmax normalization to generate attention maps $a_k$, which are then used to weight and combine the features from both streams. This adaptive weighting mechanism allows the network to selectively focus on relevant features from each component when synthesizing the final image. The weighted features are further processed through a 1$\times$1 convolution layer to produce the synthesized output. Both networks are trained end-to-end using the Adam optimizer with carefully tuned loss weights to balance the different learning objectives.

For experimental evaluation, we employed a two-stage training strategy. First, we pre-trained our model using the 721 paired samples from the MRR dataset following Algorithm~\ref{alg:training_flow}, where each pair consists of a reflection-contaminated image and its corresponding reflection-free image. Subsequently, we fine-tuned the pre-trained model using Algorithm~\ref{alg:inference_flow_updated} on the remaining 1,621 unpaired samples, alternating between updating the RRN and RSN using separate mini-batches of reflection-contaminated and reflection-free images. This sequential training approach enables the model to first learn basic reflection removal from supervised data, then adapt to domain-specific characteristics through self-supervision. For fair comparison, all baseline methods were trained using the same paired training data from the MRR dataset, and the quantitative results reported in Table~\ref{tab:comparison_results} were obtained using our final model after both training stages.

\subsection{Evaluation Metrics}

To comprehensively assess the performance of the proposed reflection removal method, we employ a combination of traditional image quality metrics and advanced perceptual quality measures. Specifically, we applies Peak Signal-to-Noise Ratio (PSNR), Structural Similarity Index (SSIM)  \cite{wang2004image}, Learned Perceptual Image Patch Similarity (LPIPS) \cite{zhang2018unreasonable} and the proposed Reflection Artifact Measure (RAM).

\begin{figure}[tbp]
	\centering
	\includegraphics[width=\linewidth]{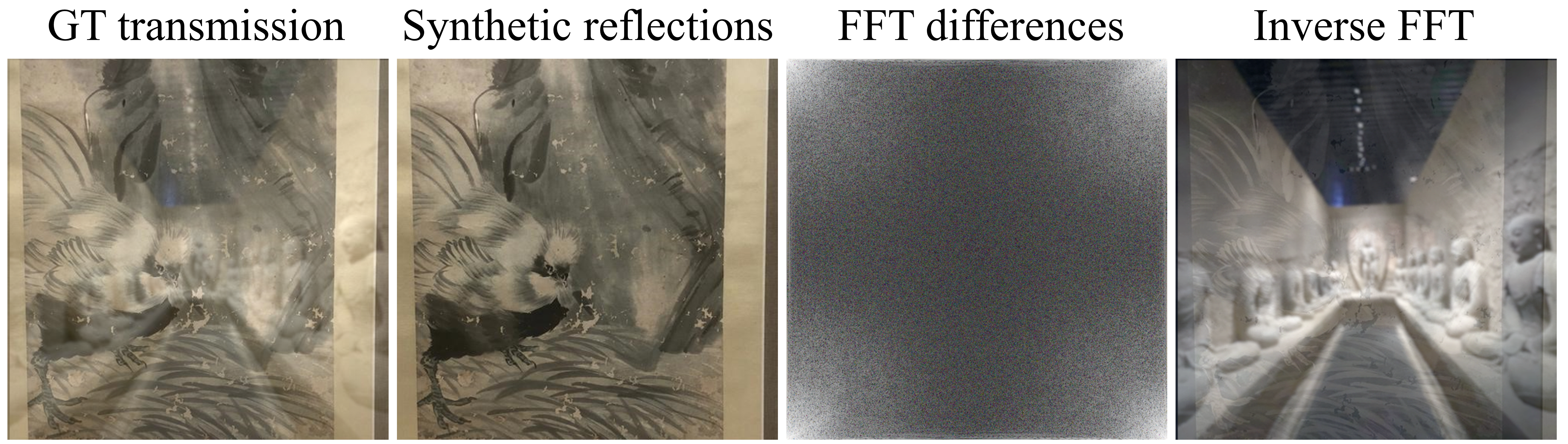}
	\caption{The Fourier transform difference proves to be an effective method for recognizing reflections, making it an ideal evaluation metric for assessing reflection removal results.}
	\label{fig:ram}
\end{figure}

The Reflection Artifact Measure (RAM) is a novel metric designed to quantify the presence of frequency-ware reflection component in the recovered transmission layer. 
As illustrated in Fig. \ref{fig:ram}, the RAM can be defined as follows:
\begin{equation}
	RAM = \frac{1}{N} \sum_{i=1}^N \frac{\left | \mathcal{F}^{-1} \left ( \mathcal{F}(\hat{T}_i) - \mathcal{F}(T_i)  \right ) \right |_1}{\left | T_i \right |_1}
\end{equation}
where $\mathcal{F}(\cdot)$ and $\mathcal{F}^{-1}(\cdot)$ denote the Fourier transform and its inverse, respectively. By measuring the energy of the frequency-ware components and normalizing it by the total energy of the predicted transmission layer, RAM provides a quantitative assessment the reflection components in the recovered image. A lower RAM score indicates better suppression of reflection artifacts and higher quality of the reflection removal results.

\begin{table}[!t]
	\centering
	\caption{Human-centered Comparison of Evaluation Metrics on Reflection Removal Assessment on SIR$^2$ Dataset}
	\label{tab:metric_comparison}
	\begin{tabular}{lccc}
		\toprule
		Metric & \begin{tabular}[c]{@{}c@{}}Human\\ Correlation\end{tabular} & \begin{tabular}[c]{@{}c@{}}Reflection\\ Detection Rate\end{tabular} & \begin{tabular}[c]{@{}c@{}}False\\ Positive Rate\end{tabular} \\
		\midrule
		PSNR & 0.76 & 68.4\% & 12.3\% \\
		SSIM & 0.79 & 71.2\% & 9.8\% \\
		LPIPS & 0.81 & 75.8\% & 8.2\% \\
		RAM (Ours) & \textbf{0.85} & \textbf{89.5\%} & \textbf{6.5\%} \\
		\bottomrule
	\end{tabular}
\end{table}

To validate RAM's effectiveness, we conducted correlation analysis between RAM scores and human perceptual evaluations on SIR$^2$ Dataset. While PSNR and SSIM show moderate correlation with human judgments (Pearson coefficients of 0.76 and 0.79 respectively) for reflection removal quality, RAM demonstrates stronger correlation (0.85) specifically for reflection artifact detection. Additionally, RAM successfully identifies residual reflections in cases where traditional metrics fail, particularly in textured regions where PSNR/SSIM may overlook subtle reflection artifacts.

\subsection{Comparison with State-of-the-Art Methods}

We conduct a comprehensive evaluation of the method against state-of-the-art techniques, including CEILNet \cite{fan2017generic}, PercepNet \cite{zhang2018single}, BDN \cite{yang2018seeing}, ERRNet \cite{wei2019single}, KimFe et al. \cite{wieschollek2018separating}, DSRN \cite{hu2023single} and SDN \cite{chang2020siamese}, on three benchmark datasets: SIR$^2$ \cite{wan2017benchmarking}, FRR\cite{lei2021robust} and the proposed MRR dataset. These datasets cover various real-world and synthetic scenarios, enabling a thorough assessment of reflection removal performance.

Table \ref{tab:comparison_results} demonstrates that the proposed method outperforms state-of-the-art reflection removal techniques on the SIR$^2$, FRR, and MRR datasets across all evaluation metrics. Deep learning approaches such as CEILNet \cite{fan2017generic}, PercepNet \cite{zhang2018single}, BDN \cite{yang2018seeing}, and ERRNet \cite{wei2019single} achieve competitive results but are surpassed by the proposed method. Our approach achieves an average PSNR gain of 0.50-3.84 dB, SSIM improvement of 0.005-0.074, LPIPS reduction of 0.003-0.057, and RAM decrease of 0.0050-0.0561 compared to the baselines. The superior quantitative results on diverse datasets, including significant improvements in the proposed RAM, demonstrate the effectiveness of combining cycle-consistency and denoising diffusion models for single image reflection removal, particularly in suppressing reflection artifacts. Specifically, the proposed method achieves RAM scores of 0.0549, 0.0437, and 0.0498 on the SIR$^2$, FRR, and MRR datasets, respectively, which are considerably lower than the other methods, indicating better suppression of reflection artifacts.

Further analysis reveals scenario-specific performance variations. Our method shows particular advantages in handling complex museum artifacts (+2.1dB PSNR improvement over ERRNet) and textured surfaces (+1.8dB over SDN), likely due to the diffusion model's strong capability in handling multimodal distributions. However, for scenes with extremely strong reflections or motion blur, the performance gain is more modest (+0.3dB over DSRN). Traditional methods like BDN perform competitively on simple flat surfaces but struggle with layered reflections where our approach excels.

\begin{table*}[!t]
	\centering
	\caption{Quantitative comparison with state-of-the-art methods on the SIR$^2$, FRR and MRR datasets.}
	\label{tab:comparison_results}
	\resizebox{\linewidth}{!}{
		\begin{tabular}{lccccccccccccc}
			\toprule
			& \multicolumn{4}{c}{SIR$^2$} & \multicolumn{4}{c}{FRR} & \multicolumn{4}{c}{MRR} & \\
			\cmidrule(lr){2-5} \cmidrule(lr){6-9} \cmidrule(lr){10-13}
			Method & PSNR & SSIM & LPIPS & RAM & PSNR & SSIM & LPIPS & RAM & PSNR & SSIM & LPIPS & RAM & \#Params\\
			\midrule
			CEILNet \cite{fan2017generic} & 25.23 & 0.861 & 0.132 & 0.1110 & 22.14 & 0.793 & 0.185 & 0.0998 & 21.37 & 0.778 & 0.193 & 0.1061 & 45.2M\\
			PercepNet \cite{zhang2018single} & 26.17 & 0.879 & 0.116 & 0.0924 & 23.08 & 0.815 & 0.169 & 0.0812 & 22.25 & 0.802 & 0.177 & 0.0873 & 58.7M\\
			BDN \cite{yang2018seeing} & 27.35 & 0.895 & 0.099 & 0.0736 & 23.92 & 0.828 & 0.157 & 0.0624 & 23.11 & 0.817 & 0.164 & 0.0685 & 64.3M\\
			ERRNet \cite{wei2019single} & 27.63 & 0.901 & 0.095 & 0.0662 & 24.21 & 0.834 & 0.151 & 0.0550 & 23.48 & 0.825 & 0.158 & 0.0611 & 71.5M\\
			KimFe et al. \cite{wieschollek2018separating} & 24.57 & 0.842 & 0.147 & 0.1048 & 21.63 & 0.776 & 0.198 & 0.0936 & 20.89 & 0.761 & 0.205 & 0.0999 & 43.8M\\
			DSRN \cite{hu2023single} & 27.82 & 0.905 & 0.092 & 0.0624 & 24.39 & 0.837 & 0.147 & 0.0512 & 23.74 & 0.830 & 0.153 & 0.0573 & 82.1M\\
			SDN \cite{chang2020siamese} & 27.91 & 0.907 & 0.090 & 0.0599 & 24.52 & 0.839 & 0.145 & 0.0487 & 23.86 & 0.832 & 0.151 & 0.0548 & 85.3M\\
			Ours & \textbf{28.41} & \textbf{0.912} & \textbf{0.087} & \textbf{0.0549} & \textbf{24.76} & \textbf{0.841} & \textbf{0.142} & \textbf{0.0437} & \textbf{24.15} & \textbf{0.835} & \textbf{0.148} & \textbf{0.0498} & 89.7M\\
			\bottomrule
		\end{tabular}
	}
\end{table*}

\begin{figure*}[tbp]
	\centering
	\includegraphics[width=\linewidth]{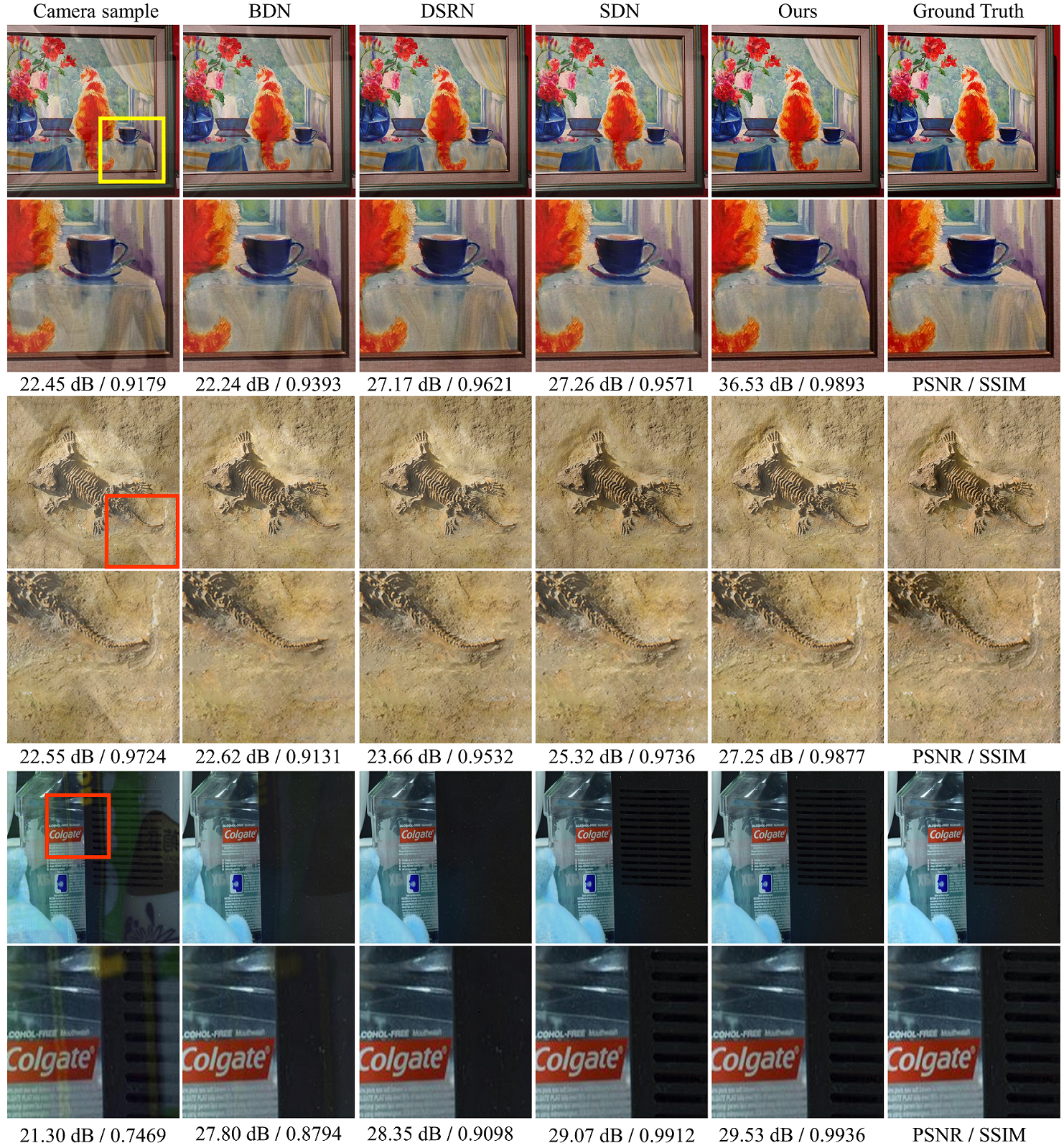}
	\caption{Qualitative comparison of the proposed method with state-of-the-art techniques on sample images from the SIR$^2$ and MRR datasets. From left to right: Input image with reflections, results from BDN \cite{yang2018seeing}, DSRN \cite{hu2023single}, SDN \cite{chang2020siamese} and the proposed method.}
	\label{fig:qualitative_comparison}
\end{figure*}

Fig. \ref{fig:qualitative_comparison} provides qualitative comparisons of the proposed method with the baseline techniques on sample images from the SIR$^2$ and MRR datasets. The proposed method effectively removes the reflective component while preserving the original image content and structures, outperforming the existing methods in terms of visual quality and realism. Training approaches based on fully supervised learning, which rely on paired samples, can also yield reasonable results. However, these methods may encounter limitations in certain scenarios, leading to the artifacts or incomplete removal in the recovered transmission images.

While our method has a slightly larger model size (89.7M parameters) compared to recent approaches like SDN (85.3M) and DSRN (82.1M), the increased capacity primarily comes from the attention mechanism in RSN and the dual-branch structure in the diffusion model, which are essential for handling complex reflection patterns. Earlier methods like CEILNet (45.2M) and KimFe (43.8M) have smaller model sizes but show limited capability in handling complex scenes. The moderate increase in model parameters (approximately 5\% compared to SDN) brings substantial performance gains across all metrics, demonstrating a favorable trade-off between model complexity and reflection removal capability.

\subsection{Ablation experiments}

To investigate the each components in the proposed method, we conduct ablation experiments on the SIR$^2$ dataset. We evaluate the performance of the proposed method under the following settings:
\begin{itemize}
	\item Full Model: The complete proposed framework with all components and loss functions.
	\item w/o TD: The proposed framework without the Transmission Discriminator and adversarial loss.
	\item w/o Attention: The proposed framework replace the attention module with the simple superposition in the Reflective Synthesis Network.  
	\item w/o RSN: The proposed framework without the Reflective Synthesis Network and cycle-consistency loss.
\end{itemize}

\begin{figure*}[h]
	\centering
	\includegraphics[width=\linewidth]{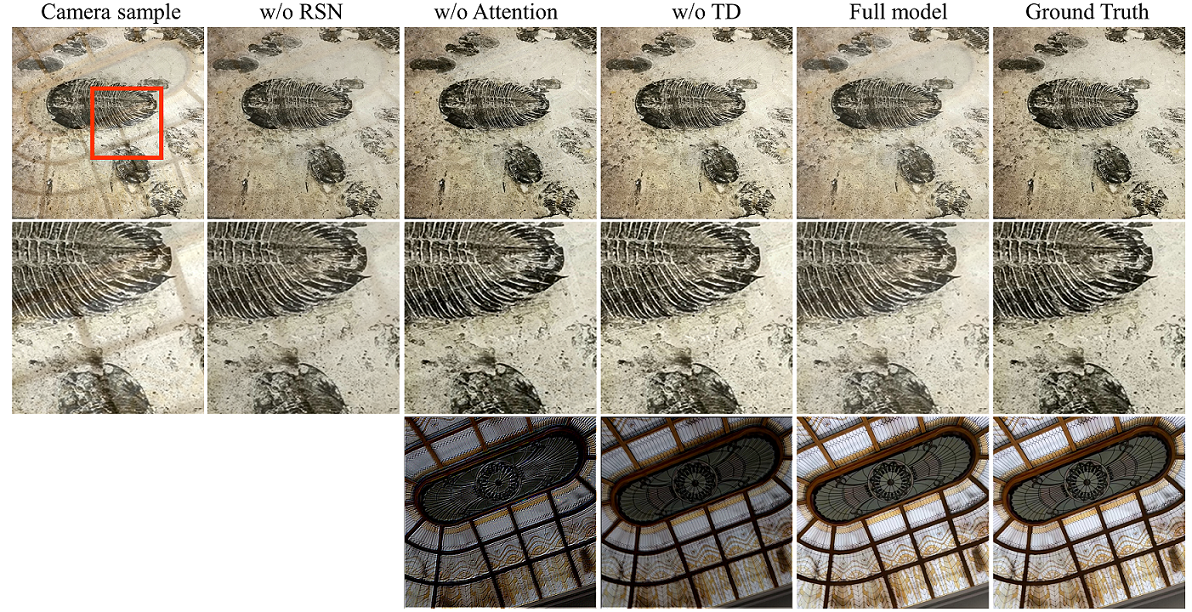}
	\caption{Qualitative ablation results on the MRR dataset. From upper to bottom: full transmission image, full model, w/o TD, w/o attention, and w/o RSN.}
	\label{fig:ablation}
\end{figure*}

Table \ref{tab:ablation_results} presents the quantitative results of the ablation study. The full model achieves the best performance across all evaluation metrics, demonstrating the effectiveness of the proposed components and loss functions. Removing the Transmission Discriminator leads to a drop in performance, with a decrease of 1.26 dB in PSNR, 0.016 in SSIM, and an increase of 0.015 in LPIPS and 0.0334 in RAM. This indicates the importance of the adversarial loss and the TD in improving the quality of the recovered transmission image.
The attention-based combination module in the RSN also contributes positively to the overall performance. Without attention module, the PSNR decreases by 2.09 dB, SSIM drops by 0.029, LPIPS increases by 0.027, and RAM increases by 0.0676, highlighting the significance of the attention mechanism in adaptively fusing the transmission and reflection components.
Removing the Reflective Synthesis Network results in the most significant performance degradation among the ablation settings. The PSNR decreases by 2.54 dB, SSIM drops by 0.041, LPIPS increases by 0.040, and RAM increases by 0.1058, emphasizing the crucial role of the RSN and the cycle-consistency loss in the proposed framework.

For computational complexity, our full model requires 89.4G FLOPs for processing a 512$\times$512 image, with the TD and RSN components contributing additional overhead compared to the baseline architectures. While removing these components reduces computational cost (65.3G FLOPs for w/o RSN), the significant performance degradation suggests that the added complexity is justified by the quality improvements. This trade-off between computational efficiency and reflection removal performance provides flexibility for different application scenarios.

Fig. \ref{fig:ablation} presents the qualitative results of the ablation study on sample images from the MRR dataset. The full model achieves the best visual quality in the recovered transmission images, successfully removing the reflections while preserving the image content. Removing the Transmission Discriminator (w/o TD) leads to incomplete removal of reflections in some regions. The absence of the attention module (w/o attention) results in artifacts and distortions in the recovered images. Without the Reflective Synthesis Network (w/o RSN), the quality of the transmission images degrades significantly, with visible residual reflections and loss of details.

\begin{table}[!t]
	\centering
	\caption{Quantitative results of the ablation study on the SIR$^2$ dataset.}
	\label{tab:ablation_results}
	\begin{tabular}{lccccc}
		\toprule
		Method & PSNR $\uparrow$ & SSIM $\uparrow$ & LPIPS $\downarrow$ & RAM $\downarrow$ & FLOPs (G) \\
		\midrule
		Full Model & \textbf{28.41} & \textbf{0.912} & \textbf{0.087} & \textbf{0.0549} & 89.4 \\
		w/o TD & 27.15 & 0.896 & 0.102 & 0.0883 & 76.2 \\
		w/o Attention & 26.32 & 0.883 & 0.114 & 0.1225 & 82.8 \\ 
		w/o RSN & 25.87 & 0.871 & 0.127 & 0.1607 & 65.3 \\
		\bottomrule
	\end{tabular}
\end{table}

\subsection{Runtime Analysis}

We analyze the computational complexity and runtime of the proposed method in comparison with existing state-of-the-art approaches. Table \ref{tab:runtime_analysis} presents the average runtime of the method with different resolutions on a single NVIDIA GeForce RTX 4090s GPU. 

\begin{table}[!t]
	\centering
	\caption{Runtime Analysis and Comparison (in seconds) on Different Image Resolutions}
	\label{tab:runtime_analysis}
	\begin{tabular}{lcccc}
		\toprule
		Method & Parameters & 256$\times$256 & 512$\times$512 & 1024$\times$1024 \\
		\midrule
		CEILNet \cite{fan2017generic} & 45.2M & 0.032 & 0.128 & 0.482 \\
		BDN \cite{yang2018seeing} & 64.3M & 0.041 & 0.156 & 0.589 \\
		ERRNet \cite{wei2019single} & 71.5M & 0.045 & 0.167 & 0.634 \\
		DSRN \cite{hu2023single} & 82.1M & 0.053 & 0.198 & 0.756 \\
		SDN \cite{chang2020siamese} & 85.3M & 0.057 & 0.212 & 0.823 \\
		Ours & 89.7M & 0.061 & 0.235 & 0.892 \\
		\bottomrule
	\end{tabular}
\end{table}

Table \ref{tab:runtime_analysis} presents the average runtime of different methods for processing images of various resolutions on a single NVIDIA GeForce RTX 4090s GPU. As expected, lighter models like CEILNet achieve faster inference times (0.128s at 512$\times$512) due to their simpler architectures. Our method shows competitive runtime performance (0.235s at 512$\times$512) despite incorporating more sophisticated components like the attention mechanism and dual-branch diffusion structure. The increased latency is justified by the significant improvement in reflection removal quality, as demonstrated in our quantitative results. All methods exhibit approximately quadratic scaling with image resolution, which aligns with the theoretical complexity of convolutional operations. For practical applications requiring real-time processing, our method can still process multiple frames per second at common resolutions while delivering superior reflection removal results.

\subsection{Real-World results}

\begin{figure}[h]
	\centering
	\includegraphics[width=0.75\linewidth]{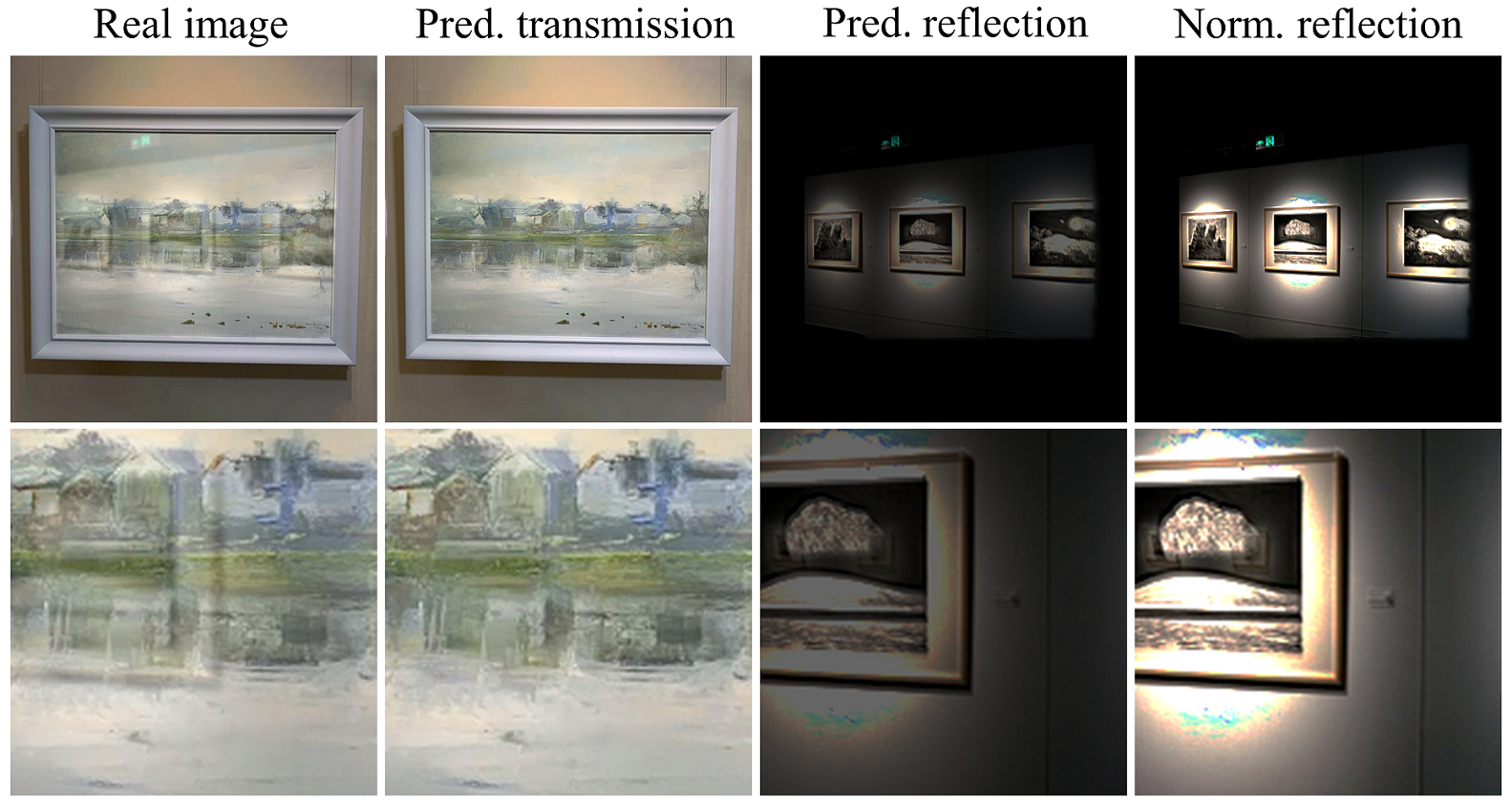}
	\caption{Reflective positions in real-world scenes are often ambiguous. Our method recovers visually accurate transmission and reflection images.}
	\label{fig:real_world_results}
\end{figure}

To further validate the effectiveness of the proposed method in practical scenarios, we evaluate its performance on real-world images with reflections. Fig. \ref{fig:real_world_results} presents a challenging case where the reflections are primarily concentrated within the painting frame, while the painting itself remains relatively reflection-free. This scenario often occurs in real-world settings, such as museums or galleries, where the reflective positions are ambiguous and can be easily confused with the actual content of the artwork.

The proposed method successfully handles this challenging case, accurately separating the reflection component from the transmission layer. The recovered transmission image preserves the intricate details and textures of the painting, while the estimated reflection image captures the reflective elements present in the frame. These results demonstrate the robustness of the proposed approach in dealing with complex real-world reflections.

\subsection{Failure Cases and Limitations}

\begin{figure}[ht]
	\centering
	\includegraphics[width=0.65\linewidth]{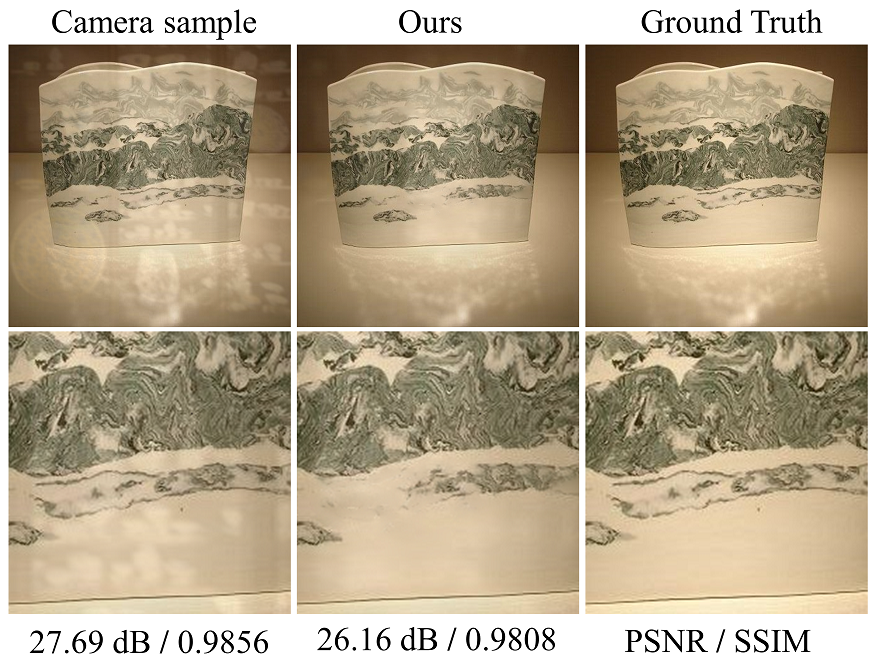}
	\caption{Failure case from the proposed method with complex pattern and reflections. Reconstructed samples achieve high metrics but misclassify details.}
	\label{fig:failure}
\end{figure}

While the proposed method demonstrates strong performance in removing reflections from single images, there are some limitations and failure cases to consider. One failure case occurs when dealing with misclassified details in complex textures, as illustrated in Fig. \ref{fig:failure}. In such scenarios, the proposed method may struggle to accurately separate the reflection component from the transmission layer, leading to over-removal of reflections in the recovered image.

The limitations in handling low-contrast reflections particularly impact applications in cultural heritage digitization and museum photography, where subtle reflections from protective glass can affect the digital archiving quality. When reflection patterns closely match the underlying object's texture or when multiple reflections overlap with varying intensities, our method may struggle to correctly separate the components. Future improvements could explore incorporating physical reflection formation models, multi-scale attention mechanisms for better feature discrimination, or auxiliary depth information to guide separation. Additionally, domain-specific pre-training on particular types of artifacts (e.g., paintings, sculptures) might help the model better handle characteristic reflection patterns in specialized settings.

\section{Conclusion}
\label{sec:conclusion}

This paper presents a self-supervised approach for single image reflection removal that combines cycle-consistency and denoising diffusion probabilistic models. The proposed method introduces a Reflective Removal Network that utilizes DDPMs to model the decomposition process and recover the transmission image, and a Reflective Synthesis Network that re-synthesizes the input using the separated components through a nonlinear attention-based mechanism. The RRN effectively handles complex reflections by leveraging the power of DDPMs, while the RSN ensures accurate reconstruction of the input image. We conduct extensive experiments on both synthetic and real-world datasets, demonstrating that the proposed technique outperforms state-of-the-art methods in terms of quantitative metrics and visual quality.   
The recovered reflection-free images exhibit high fidelity and preserve important details.
Despite existing limitations, such as incomplete removal of low-contrast reflections and reliance on synthetic training data, this work represents a significant advance in single image reflection removal, offering substantial benefits for image processing applications.

\section* {Acknowledgments}
This work is funded by National Social Science Fund of China Major Project in Artistic Studies (No.22ZD18), China Postdoctoral Science Foundation (No.2023M741411), Postdoctoral Fellowship Program of CPSF (No.GZC20240608), and Jiangsu Funding Program for Excellent Postdoctoral Talent (No.2024ZB488).

\bibliography{sn-bibliography}   
\bibliographystyle{IEEEtran}

\end{document}